%% file: PaperForReview.tex
\documentclass[10pt,twocolumn,letterpaper]{article}

\usepackage{cvpr}              

\usepackage{graphicx}
\usepackage{amsmath}
\usepackage{amssymb}
\usepackage{booktabs}
\usepackage{siunitx}
\DeclareSIUnit\px{px}

\makeatletter
\@namedef{ver@everyshi.sty}{}
\makeatother

\usepackage{todonotes}
\usepackage{enumitem}
\usepackage{amsmath}
\usepackage{comment}
\usepackage{subfig}
\usepackage{multirow}
\usepackage{pgfplots}
\usepackage{tikz}
	\usetikzlibrary{calc}
	\usetikzlibrary{positioning}
	\usetikzlibrary{shapes.geometric, arrows}

\usepackage[pagebackref,breaklinks,colorlinks]{hyperref}

\usepackage[capitalize]{cleveref}
\crefname{section}{Sec.}{Secs.}
\Crefname{section}{Section}{Sections}
\Crefname{table}{Table}{Tables}
\crefname{table}{Tab.}{Tabs.}

\def\cvprPaperID{*****} 
\def\confName{UNCV-W}
\def\confYear{2022}

\begin{document}

\title{Identifying Out-of-Distribution Samples in Real-Time for Safety-Critical 2D Object Detection with Margin Entropy Loss}

\newcommand{\thanksmsg}{
	This work has been part of a master thesis supervised by Nazih Mechbal and Prof. Dr.-Ing. Carsten Proppe.
}

\author{Yannik Blei\textsuperscript{1,2,}\thanks{\thanksmsg} \qquad Nicolas Jourdan\textsuperscript{1} \qquad Nils G\"ahlert\textsuperscript{1}\\
\textsuperscript{1}Wingcopter GmbH \qquad \textsuperscript{2}Arts et M\'etiers ParisTech\\
{\tt\small \{blei, jourdan, gaehlert\}@wingcopter.com}
}
\maketitle

\begin{abstract}
	Convolutional Neural Networks (CNNs) are nowadays often employed in vision-based perception stacks for safety-critical applications such as autonomous driving or Unmanned Aerial Vehicles (UAVs).
	Due to the safety requirements in those use cases, it is important to know the limitations of the CNN and, thus, to detect Out-of-Distribution (OOD) samples.
	In this work, we present an approach to enable OOD detection for 2D object detection by employing the margin entropy (ME) loss.
	The proposed method is easy to implement and can be applied to most existing object detection architectures.
	In addition, we introduce Separability as a metric for detecting OOD samples in object detection.
	We show that a CNN trained with the ME loss significantly outperforms OOD detection using standard confidence scores.
	At the same time, the runtime of the underlying object detection framework remains constant rendering the ME loss a powerful tool to enable OOD detection.
\end{abstract}

\input{histograms}
\input{Chapters/1_intro}

\input{Chapters/2_related_work}
\input{Chapters/3_model}

\input{Chapters/4_experiments}
\input{Chapters/5_conclusion}

{\small
	\bibliographystyle{ieee_fullname}
	\bibliography{egbib}
}

\end{document}

%% file: histograms.tex
\newcommand{\varBaselineSep}{0.213\:}
\newcommand{\varBaselineIdOod}{0.87\:}
\newcommand{\varBaselineOodBack}{0.35\:}
\newcommand{\varEnsemblesSep}{0.172\:}
\newcommand{\varEnsemblesIdOod}{0.301\:}
\newcommand{\varEnsemblesOodBack}{0.04\:}
\newcommand{\varMElossSep}{0.486\:}
\newcommand{\varMElossIdOod}{0.42\:}
\newcommand{\varMElossOodBack}{0.39\:}
\newcommand{\varMElossApNinetyFive}{0.248\:}
\newcommand{\varMElossApFifty}{0.356\:}
\newcommand{\vartwoDBasicSep}{0.172\:}
\newcommand{\vartwoDBasicIdOod}{0.301\:}
\newcommand{\vartwoDBasicOodBack}{0.04\:}
\newcommand{\varSepImprovement}{128.2\:}
\newcommand{\tikzBaselinePDF}{
    \begin{tikzpicture}

        \definecolor{color0}{rgb}{0.12156862745098,0.466666666666667,0.705882352941177}
        \definecolor{color1}{rgb}{1,0.498039215686275,0.0549019607843137}
        \definecolor{color2}{rgb}{0.172549019607843,0.627450980392157,0.172549019607843}

        \begin{axis}[
                height=0.21\textwidth,
                legend cell align={left},
                legend style={fill opacity=0.8, draw opacity=1, text opacity=1, draw=white!80!black, nodes={scale=0.75, transform shape}},
                tick pos=left,
                width=0.48\textwidth,
                x grid style={white!69.0196078431373!black},
                xmin=0, xmax=1,
                xtick style={color=black},
                y grid style={white!69.0196078431373!black},
                ylabel={Density [\si{\percent}]},
                ylabel style={yshift=-10pt},
                ymin=-1.17296862429606, ymax=24.6323411102172,
                ytick style={color=black}
            ]
            \addplot [semithick, color0]
            table [row sep=\\] {%
                    0.0025 1.2018779342723\\0.0275 0.275430359937402\\0.0525 0.200312989045383\\0.0775 0.162754303599374\\0.1025 0.200312989045383\\0.1275 0.137715179968701\\0.1525 0.112676056338028\\0.1775 0.125195618153365\\0.2025 0.137715179968701\\0.2275 0.100156494522692\\0.2525 0.137715179968701\\0.2775 0.225352112676056\\0.3025 0.125195618153365\\0.3275 0.187793427230047\\0.3525 0.112676056338028\\0.3775 0.100156494522692\\0.4025 0.125195618153365\\0.4275 0.112676056338028\\0.4525 0.125195618153365\\0.4775 0.137715179968701\\0.5025 0.17527386541471\\0.5275 0.162754303599374\\0.5525 0.187793427230047\\0.5775 0.137715179968701\\0.6025 0.25039123630673\\0.6275 0.17527386541471\\0.6525 0.150234741784037\\0.6775 0.187793427230047\\0.7025 0.187793427230047\\0.7275 0.350547730829422\\0.7525 0.150234741784037\\0.7775 0.350547730829421\\0.8025 0.413145539906103\\0.8275 0.851330203442879\\0.8525 2.3661971830986\\0.8775 10.3787167449139\\0.9025 8.83881064162754\\0.9275 6.61032863849765\\0.9525 3.51799687010954\\0.9775 0.513302034428797\\};
            \addlegendentry{ID FG}
            \addplot [semithick, color1]
            table [row sep=\\] {%
                    0.0025 4.33734939759036\\0.0275 0.72289156626506\\0.0525 0.963855421686747\\0.0775 0.602409638554217\\0.1025 0.481927710843374\\0.1275 0.240963855421687\\0.1525 0.36144578313253\\0.1775 0.36144578313253\\0.2025 0.36144578313253\\0.2275 0.481927710843374\\0.2525 0.36144578313253\\0.2775 0.240963855421687\\0.3025 0.240963855421687\\0.3275 0.240963855421687\\0.3525 0.240963855421687\\0.3775 0.72289156626506\\0.4025 0.481927710843373\\0.4275 0.602409638554218\\0.4525 0.36144578313253\\0.4775 0.722891566265061\\0.5025 0.120481927710843\\0.5275 0.120481927710843\\0.5525 0.120481927710843\\0.5775 0.481927710843373\\0.6025 0.96385542168675\\0.6275 0.963855421686746\\0.6525 0.481927710843373\\0.6775 0.36144578313253\\0.7025 0.481927710843373\\0.7275 1.44578313253013\\0.7525 0.602409638554216\\0.7775 0.481927710843373\\0.8025 1.56626506024096\\0.8275 2.7710843373494\\0.8525 3.73493975903616\\0.8775 4.57831325301204\\0.9025 3.49397590361445\\0.9275 3.01204819277108\\0.9525 1.08433734939759\\0.9775 0\\};
            \addlegendentry{OOD}
            \addplot [semithick, color2]
            table [row sep=\\] {%
                    0.0025 23.4593724859212\\0.0275 1.81818181818182\\0.0525 1.02976669348351\\0.0775 0.901045856798069\\0.1025 0.707964601769912\\0.1275 0.466613032984714\\0.1525 0.498793242156074\\0.1775 0.386162510056315\\0.2025 0.482703137570394\\0.2275 0.289621882542237\\0.2525 0.370072405470635\\0.2775 0.241351568785197\\0.3025 0.321802091713597\\0.3275 0.176991150442478\\0.3525 0.193081255028158\\0.3775 0.241351568785197\\0.4025 0.193081255028158\\0.4275 0.176991150442478\\0.4525 0.160901045856798\\0.4775 0.273531777956557\\0.5025 0.386162510056315\\0.5275 0.160901045856798\\0.5525 0.225261464199517\\0.5775 0.209171359613837\\0.6025 0.144810941271119\\0.6275 0.353982300884955\\0.6525 0.434432823813354\\0.6775 0.160901045856798\\0.7025 0.305711987127916\\0.7275 0.402252614641997\\0.7525 0.289621882542236\\0.7775 0.482703137570394\\0.8025 0.209171359613837\\0.8275 0.418342719227675\\0.8525 0.884955752212393\\0.8775 1.22284794851166\\0.9025 0.820595333869669\\0.9275 0.337892196299276\\0.9525 0.144810941271118\\0.9775 0.0160901045856799\\};
            \addlegendentry{ID BG}
        \end{axis}

    \end{tikzpicture}
}
\newcommand{\tikzEnsemblesPDF}{
    \begin{tikzpicture}

        \definecolor{color0}{rgb}{0.12156862745098,0.466666666666667,0.705882352941177}
        \definecolor{color1}{rgb}{1,0.498039215686275,0.0549019607843137}
        \definecolor{color2}{rgb}{0.172549019607843,0.627450980392157,0.172549019607843}

        \begin{axis}[
                height=0.4\textwidth,
                legend cell align={left},
                legend style={fill opacity=0.8, draw opacity=1, text opacity=1, draw=white!80!black, nodes={scale=0.75, transform shape}},
                tick pos=left,
                width=0.48\textwidth,
                x grid style={white!69.0196078431373!black},
                xlabel={Confidence},
                xmin=0, xmax=1,
                xtick style={color=black},
                y grid style={white!69.0196078431373!black},
                ylabel={Density [\si{\percent}]},
                ymin=-1.27380952380952, ymax=26.75,
                ytick style={color=black}
            ]
            \addplot [semithick, color0]
            table [row sep=\\] {%
                    0.0025 1.42724170027924\\0.0275 0.297859137449581\\0.0525 0.248215947874651\\0.0775 0.198572758299721\\0.1025 0.260626745268384\\0.1275 0.173751163512256\\0.1525 0.136518771331058\\0.1775 0.223394353087186\\0.2025 0.124107973937326\\0.2275 0.136518771331058\\0.2525 0.173751163512256\\0.2775 0.173751163512256\\0.3025 0.111697176543593\\0.3275 0.136518771331058\\0.3525 0.124107973937326\\0.3775 0.161340366118523\\0.4025 0.0620539869686627\\0.4275 0.148929568724791\\0.4525 0.0620539869686627\\0.4775 0.136518771331058\\0.5025 0.136518771331058\\0.5275 0.14892956872479\\0.5525 0.111697176543593\\0.5775 0.124107973937325\\0.6025 0.173751163512256\\0.6275 0.136518771331058\\0.6525 0.161340366118523\\0.6775 0.260626745268383\\0.7025 0.161340366118523\\0.7275 0.310269934843315\\0.7525 0.260626745268383\\0.7775 0.359913124418244\\0.8025 0.533664287930499\\0.8275 0.93080980452994\\0.8525 3.67359602854485\\0.8775 9.71765435929258\\0.9025 8.05460750853242\\0.9275 5.33664287930499\\0.9525 3.6984176233323\\0.9775 1.19143654979833\\};
            \addlegendentry{ID FG}
            \addplot [semithick, color1]
            table [row sep=\\] {%
                    0.0025 2.38095238095238\\0.0275 1.19047619047619\\0.0525 0.634920634920635\\0.0775 0.873015873015873\\0.1025 0.476190476190476\\0.1275 0.873015873015872\\0.1525 0.555555555555556\\0.1775 0.476190476190476\\0.2025 0.634920634920635\\0.2275 0.634920634920635\\0.2525 0.396825396825396\\0.2775 1.03174603174603\\0.3025 1.03174603174603\\0.3275 1.74603174603174\\0.3525 6.90476190476191\\0.3775 10.5555555555555\\0.4025 3.09523809523809\\0.4275 0.714285714285715\\0.4525 0.0793650793650793\\0.4775 0\\0.5025 0\\0.5275 0.158730158730159\\0.5525 0.238095238095238\\0.5775 0.0793650793650793\\0.6025 0.158730158730159\\0.6275 0.158730158730159\\0.6525 0.238095238095238\\0.6775 0.238095238095238\\0.7025 0.0793650793650793\\0.7275 0.476190476190478\\0.7525 0.0793650793650793\\0.7775 0.396825396825396\\0.8025 0.634920634920634\\0.8275 0.396825396825396\\0.8525 0.873015873015876\\0.8775 0.317460317460317\\0.9025 0.714285714285714\\0.9275 0.317460317460317\\0.9525 0.158730158730159\\0.9775 0\\};
            \addlegendentry{OOD}
            \addplot [semithick, color2]
            table [row sep=\\] {%
                    0.0025 25.4761904761905\\0.0275 2.42857142857143\\0.0525 1.13095238095238\\0.0775 0.857142857142857\\0.1025 0.55952380952381\\0.1275 0.464285714285714\\0.1525 0.345238095238095\\0.1775 0.333333333333333\\0.2025 0.321428571428572\\0.2275 0.440476190476191\\0.2525 0.345238095238095\\0.2775 0.309523809523809\\0.3025 0.166666666666667\\0.3275 0.214285714285714\\0.3525 0.25\\0.3775 0.273809523809524\\0.4025 0.25\\0.4275 0.273809523809524\\0.4525 0.238095238095238\\0.4775 0.166666666666667\\0.5025 0.19047619047619\\0.5275 0.25\\0.5525 0.154761904761905\\0.5775 0.19047619047619\\0.6025 0.166666666666667\\0.6275 0.154761904761905\\0.6525 0.226190476190476\\0.6775 0.297619047619047\\0.7025 0.178571428571428\\0.7275 0.214285714285715\\0.7525 0.142857142857143\\0.7775 0.25\\0.8025 0.297619047619047\\0.8275 0.19047619047619\\0.8525 0.678571428571431\\0.8775 0.797619047619047\\0.9025 0.392857142857143\\0.9275 0.154761904761905\\0.9525 0.226190476190476\\0.9775 0\\};
            \addlegendentry{ID BG}
        \end{axis}

    \end{tikzpicture}
}
\newcommand{\tikzMElossPDF}{
    \begin{tikzpicture}

        \definecolor{color0}{rgb}{0.12156862745098,0.466666666666667,0.705882352941177}
        \definecolor{color1}{rgb}{1,0.498039215686275,0.0549019607843137}
        \definecolor{color2}{rgb}{0.172549019607843,0.627450980392157,0.172549019607843}

        \begin{axis}[
                height=0.21\textwidth,
                legend cell align={left},
                legend style={fill opacity=0.8, draw opacity=1, text opacity=1, draw=white!80!black, nodes={scale=0.75, transform shape}},
                tick pos=left,
                width=0.48\textwidth,
                x grid style={white!69.0196078431373!black},
                xlabel={Confidence},
                xmin=0, xmax=1,
                xtick style={color=black},
                y grid style={white!69.0196078431373!black},
                ylabel={Density [\si{\percent}]},
                ylabel style={yshift=-10pt},
                ymin=-1.53882653526782, ymax=32.3153572406241,
                ytick style={color=black}
            ]
            \addplot [semithick, color0]
            table [row sep=\\] {%
                    0.0025 2.59563615755171\\0.0275 0.612071408330972\\0.0525 0.408047605553981\\0.0775 0.453386228393313\\0.1025 0.487390195522811\\0.1275 0.476055539812978\\0.1525 0.328705015585152\\0.1775 0.260697081326155\\0.2025 0.170019835647492\\0.2275 0.28336639274582\\0.2525 0.147350524227826\\0.2775 0.226693114196656\\0.3025 0.215358458486824\\0.3275 0.124681212808161\\0.3525 0.0793425899688298\\0.3775 0.147350524227826\\0.4025 0.136015868517994\\0.4275 0.170019835647492\\0.4525 0.102011901388495\\0.4775 0.147350524227827\\0.5025 0.20402380277699\\0.5275 0.136015868517994\\0.5525 0.147350524227826\\0.5775 0.136015868517994\\0.6025 0.181354491357326\\0.6275 0.215358458486823\\0.6525 0.170019835647492\\0.6775 0.249362425616322\\0.7025 0.351374327004817\\0.7275 0.532728818362144\\0.7525 0.974780391045621\\0.7775 1.60952111079626\\0.8025 3.77444035137432\\0.8275 5.92802493624256\\0.8525 7.40153017852085\\0.8775 4.48852366109379\\0.9025 2.7996599603287\\0.9275 2.18758855199773\\0.9525 0.895437801076791\\0.9775 0.0453386228393314\\};
            \addlegendentry{ID FG}
            \addplot [semithick, color1]
            table [row sep=\\] {%
                    0.0025 0\\0.0275 1.35593220338983\\0.0525 0.677966101694915\\0.0775 0.677966101694915\\0.1025 0\\0.1275 0.677966101694915\\0.1525 0\\0.1775 0\\0.2025 0.677966101694915\\0.2275 0\\0.2525 0.677966101694915\\0.2775 1.35593220338983\\0.3025 2.03389830508475\\0.3275 1.35593220338983\\0.3525 3.38983050847458\\0.3775 10.1694915254237\\0.4025 16.271186440678\\0.4275 0.677966101694916\\0.4525 0\\0.4775 0\\0.5025 0\\0.5275 0\\0.5525 0\\0.5775 0\\0.6025 0\\0.6275 0\\0.6525 0\\0.6775 0\\0.7025 0\\0.7275 0\\0.7525 0\\0.7775 0\\0.8025 0\\0.8275 0\\0.8525 0\\0.8775 0\\0.9025 0\\0.9275 0\\0.9525 0\\0.9775 0\\};
            \addlegendentry{OOD}
            \addplot [semithick, color2]
            table [row sep=\\] {%
                    0.0025 30.7765307053563\\0.0275 2.52577789944338\\0.0525 1.18258965234054\\0.0775 0.722693676430331\\0.1025 0.565745049730815\\0.1275 0.386896614654621\\0.1525 0.324847157587371\\0.1775 0.310247285336253\\0.2025 0.208048179578429\\0.2275 0.215348115703988\\0.2525 0.259147732457341\\0.2775 0.233597956017885\\0.3025 0.182498403138973\\0.3275 0.182498403138972\\0.3525 0.211698147641208\\0.3775 0.105849073820604\\0.4025 0.0875992335067067\\0.4275 0.0474495848161329\\0.4525 0.0693493931928095\\0.4775 0.0620494570672507\\0.5025 0.0547495209416917\\0.5275 0.0547495209416917\\0.5525 0.0547495209416917\\0.5775 0.0510995528789123\\0.6025 0.0693493931928098\\0.6275 0.0547495209416917\\0.6525 0.0547495209416917\\0.6775 0.0693493931928095\\0.7025 0.0510995528789123\\0.7275 0.0583994890044714\\0.7525 0.124098914134501\\0.7775 0.149648690573957\\0.8025 0.175198467013413\\0.8275 0.120448946071722\\0.8525 0.0948991696322661\\0.8775 0.0437996167533534\\0.9025 0.0364996806277945\\0.9275 0.0218998083766767\\0.9525 0\\0.9775 0\\};
            \addlegendentry{ID BG}
        \end{axis}

    \end{tikzpicture}
}
\newcommand{\tikztwoDBasicPDF}{
    \begin{tikzpicture}

        \definecolor{color0}{rgb}{0.12156862745098,0.466666666666667,0.705882352941177}
        \definecolor{color1}{rgb}{1,0.498039215686275,0.0549019607843137}
        \definecolor{color2}{rgb}{0.172549019607843,0.627450980392157,0.172549019607843}

        \begin{axis}[
                height=0.4\textwidth,
                legend cell align={left},
                legend style={fill opacity=0.8, draw opacity=1, text opacity=1, draw=white!80!black, nodes={scale=0.75, transform shape}},
                tick pos=left,
                width=0.48\textwidth,
                x grid style={white!69.0196078431373!black},
                xlabel={Confidence},
                xmin=0, xmax=1,
                xtick style={color=black},
                y grid style={white!69.0196078431373!black},
                ylabel={Probability},
                ymin=-1.27380952380952, ymax=26.75,
                ytick style={color=black}
            ]
            \addplot [semithick, color0]
            table [row sep=\\] {%
                    0.0025 1.42724170027924\\0.0275 0.297859137449581\\0.0525 0.248215947874651\\0.0775 0.198572758299721\\0.1025 0.260626745268384\\0.1275 0.173751163512256\\0.1525 0.136518771331058\\0.1775 0.223394353087186\\0.2025 0.124107973937326\\0.2275 0.136518771331058\\0.2525 0.173751163512256\\0.2775 0.173751163512256\\0.3025 0.111697176543593\\0.3275 0.136518771331058\\0.3525 0.124107973937326\\0.3775 0.161340366118523\\0.4025 0.0620539869686627\\0.4275 0.148929568724791\\0.4525 0.0620539869686627\\0.4775 0.136518771331058\\0.5025 0.136518771331058\\0.5275 0.14892956872479\\0.5525 0.111697176543593\\0.5775 0.124107973937325\\0.6025 0.173751163512256\\0.6275 0.136518771331058\\0.6525 0.161340366118523\\0.6775 0.260626745268383\\0.7025 0.161340366118523\\0.7275 0.310269934843315\\0.7525 0.260626745268383\\0.7775 0.359913124418244\\0.8025 0.533664287930499\\0.8275 0.93080980452994\\0.8525 3.67359602854485\\0.8775 9.71765435929258\\0.9025 8.05460750853242\\0.9275 5.33664287930499\\0.9525 3.6984176233323\\0.9775 1.19143654979833\\};
            \addlegendentry{ID FG}
            \addplot [semithick, color1]
            table [row sep=\\] {%
                    0.0025 2.38095238095238\\0.0275 1.19047619047619\\0.0525 0.634920634920635\\0.0775 0.873015873015873\\0.1025 0.476190476190476\\0.1275 0.873015873015872\\0.1525 0.555555555555556\\0.1775 0.476190476190476\\0.2025 0.634920634920635\\0.2275 0.634920634920635\\0.2525 0.396825396825396\\0.2775 1.03174603174603\\0.3025 1.03174603174603\\0.3275 1.74603174603174\\0.3525 6.90476190476191\\0.3775 10.5555555555555\\0.4025 3.09523809523809\\0.4275 0.714285714285715\\0.4525 0.0793650793650793\\0.4775 0\\0.5025 0\\0.5275 0.158730158730159\\0.5525 0.238095238095238\\0.5775 0.0793650793650793\\0.6025 0.158730158730159\\0.6275 0.158730158730159\\0.6525 0.238095238095238\\0.6775 0.238095238095238\\0.7025 0.0793650793650793\\0.7275 0.476190476190478\\0.7525 0.0793650793650793\\0.7775 0.396825396825396\\0.8025 0.634920634920634\\0.8275 0.396825396825396\\0.8525 0.873015873015876\\0.8775 0.317460317460317\\0.9025 0.714285714285714\\0.9275 0.317460317460317\\0.9525 0.158730158730159\\0.9775 0\\};
            \addlegendentry{OOD}
            \addplot [semithick, color2]
            table [row sep=\\] {%
                    0.0025 25.4761904761905\\0.0275 2.42857142857143\\0.0525 1.13095238095238\\0.0775 0.857142857142857\\0.1025 0.55952380952381\\0.1275 0.464285714285714\\0.1525 0.345238095238095\\0.1775 0.333333333333333\\0.2025 0.321428571428572\\0.2275 0.440476190476191\\0.2525 0.345238095238095\\0.2775 0.309523809523809\\0.3025 0.166666666666667\\0.3275 0.214285714285714\\0.3525 0.25\\0.3775 0.273809523809524\\0.4025 0.25\\0.4275 0.273809523809524\\0.4525 0.238095238095238\\0.4775 0.166666666666667\\0.5025 0.19047619047619\\0.5275 0.25\\0.5525 0.154761904761905\\0.5775 0.19047619047619\\0.6025 0.166666666666667\\0.6275 0.154761904761905\\0.6525 0.226190476190476\\0.6775 0.297619047619047\\0.7025 0.178571428571428\\0.7275 0.214285714285715\\0.7525 0.142857142857143\\0.7775 0.25\\0.8025 0.297619047619047\\0.8275 0.19047619047619\\0.8525 0.678571428571431\\0.8775 0.797619047619047\\0.9025 0.392857142857143\\0.9275 0.154761904761905\\0.9525 0.226190476190476\\0.9775 0\\};
            \addlegendentry{ID BG}
        \end{axis}

    \end{tikzpicture}
}
\newcommand{\tabBaseline}{\begin{tabular}{rrrr}
        \hline
        separability & ood\_back\_thres & id\_ood\_thres & ap50 \\
        \hline
        0.213        & 0.35             & 0.87           & nan  \\
        \hline
    \end{tabular}}
\newcommand{\tabEnsembles}{\begin{tabular}{rrrr}
        \hline
        separability & ood\_back\_thres & id\_ood\_thres & ap50 \\
        \hline
        0.172        & 0.04             & 0.301          & nan  \\
        \hline
    \end{tabular}}
\newcommand{\tabMEloss}{\begin{tabular}{rrrr}
        \hline
        separability & ood\_back\_thres & id\_ood\_thres & ap50  \\
        \hline
        0.486        & 0.39             & 0.42           & 0.356 \\
        \hline
    \end{tabular}}
\newcommand{\tabtwoDBasic}{\begin{tabular}{rrrr}
        \hline
        separability & ood\_back\_thres & id\_ood\_thres & ap50 \\
        \hline
        0.172        & 0.04             & 0.301          & nan  \\
        \hline
    \end{tabular}}
\newcommand{\tabGlobal}{\begin{tabular}{lrr}
        \hline
        Method   & Separability & ap50  \\
        \hline
        Baseline & 0.213        & nan   \\
        ME Loss  & 0.486        & 0.356 \\
        \hline
    \end{tabular}}

%% file: Chapters/1_intro.tex
\section{Introduction}
The development of UAVs strongly accelerated within the last decade through advances in robotics and artificial intelligence.
In order to mitigate collision risks, Detect-and-Avoid (DAA) systems are being developed to provide safe autonomy to UAVs by staying \emph{well clear} from other airborne traffic \cite{astmf3442}.
%
Similarly to perception stacks of other autonomous vehicles, computer vision has emerged as a mode of detecting and tracking other airspace participants.

\begin{figure}[t]
    \centering
    \includegraphics[width=0.29\columnwidth]{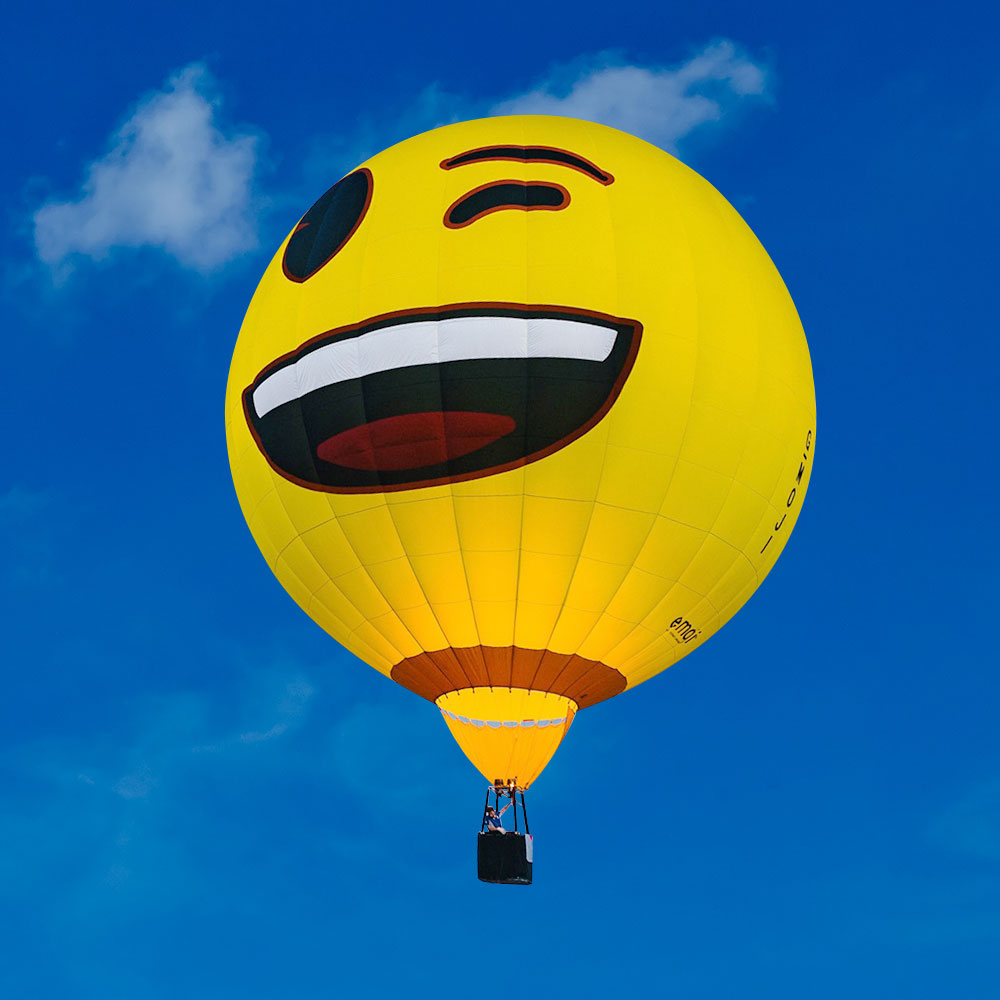}
    \includegraphics[width=0.29\columnwidth]{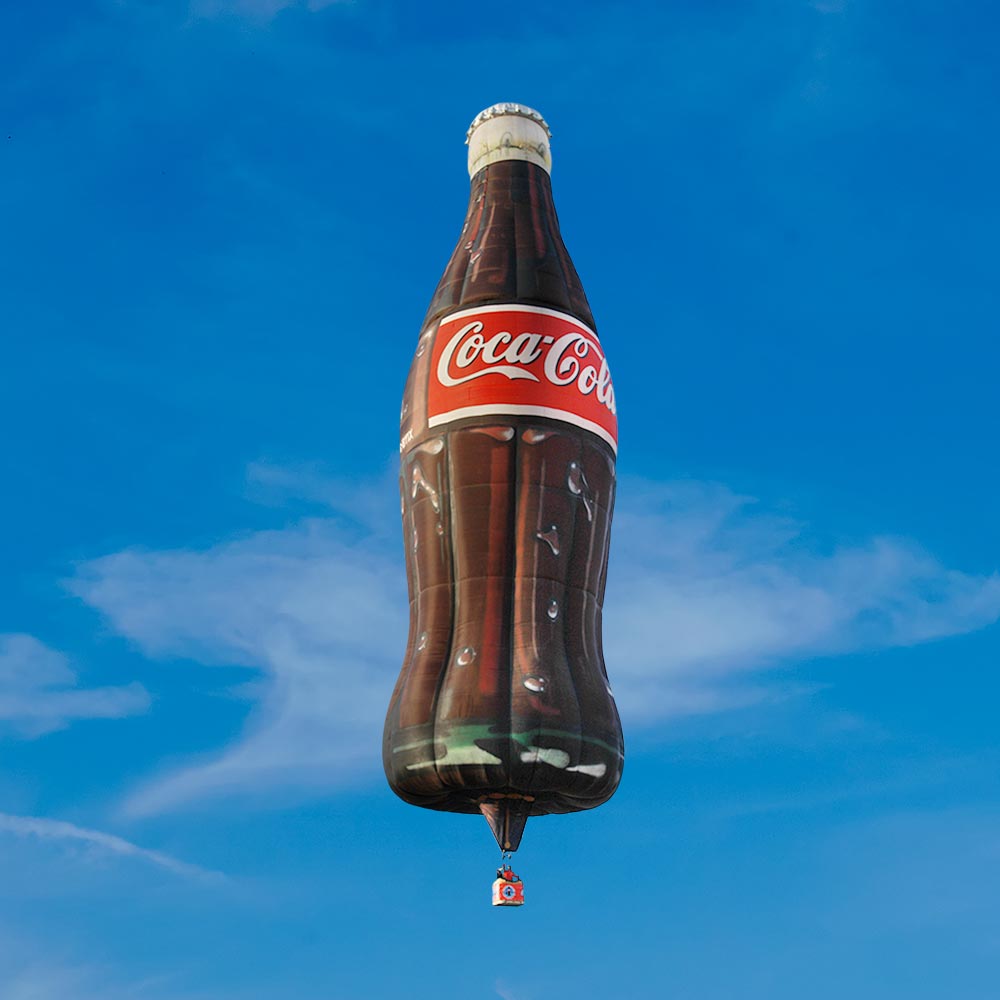}
    \includegraphics[width=0.29\columnwidth]{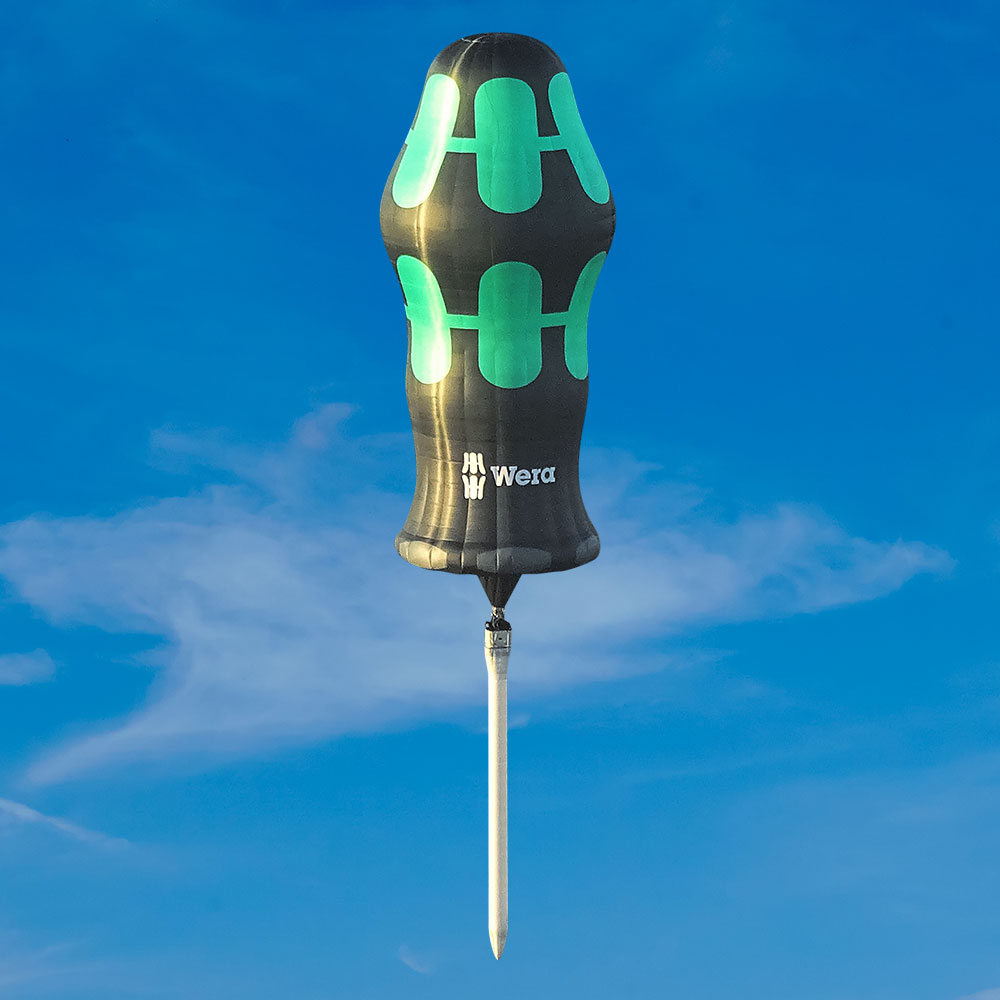}
    \caption{Exemplary OOD hot air balloons.
        Most hot air balloons have a characteristic shape, while others have a rather creative design \cite{hot_air_ood}.
    }
    \label{fig:balloons}
\end{figure}

CNNs have achieved state-of-the-art performance in computer vision tasks -- \eg semantic segmentation and object detection -- and have thus become the foundation of modern environment perception systems.
In the use case of detecting and avoiding other objects, vision-based DAA systems often employ 2D object detection CNN architectures such as Yolo \cite{redmon2016you} or SSD \cite{liu2016ssd}.
While neural networks in general have been deployed with great success, they typically do not provide self-evaluation with respect to their predictions \cite{jourdan_identication_2020}.
DAA systems are safety-critical components of UAVs, and it is, therefore, necessary that they actively self-identify their limitations as there may be no human pilot available as a fallback option \cite{CoDANN2,easa_ai_level1}.
Limitations include scene configurations or object classes that were not part of the training data used to train the neural networks.
This type of data is often referred to as \emph{Out-of-Distribution} (OOD) data in contrast to \emph{In-Distribution} (ID) data used during the training stage.
In the presence of OOD data, the desired behavior of the object detection system would be to express low confidence or high uncertainty for the detection but still identify relevant objects in contrast to the background.
An example for possible OOD data is shown in \autoref{fig:balloons}.
The DAA system, however, must perceive all of those objects and initiate countermeasures to avoid potential collisions.

Methods for uncertainty estimation and detection of OOD instances in CNNs often rely on sampling the network multiple times \eg, \cite{gal_dropout_2016,lakshminarayanan2017simple}, which is infeasible in computationally constrained environments such as small UAVs.
Thus, we present and evaluate a simple yet effective training strategy that can extend most existing object detection architectures with uncertainty estimation for OOD detection.
We exemplary extend the YoloX \cite{ge2021yolox} framework and evaluate the performance of the proposed method on a semi-synthetic aerial dataset.
We show that the OOD performance significantly exceeds the baseline performance while keeping the runtime of the detection system unchanged.

\bigbreak

%% file: Chapters/2_related_work.tex
\section{Related Work}
Out-of-distribution detection is a popular research stream for various computer vision tasks, such as image classification \cite{hendrycks_baseline_2018,lee_simple_2018,liang_enhancing_2020} and semantic segmentation \cite{chan2021entropy,kendall2015bayesian,jourdan_identication_2020}.
More recently, the topic is gaining traction in object detection research as well \cite{dhamija2020overlooked}.
Concerning the overall methodology and runtime, we divide related OOD approaches for object detection into \emph{sampling} methods and \emph{sampling free} methods.

Sampling methods often rely on test-time activation of dropout \cite{srivastava2014dropout} layers to approximate inference of a Bayesian Neural Network (BNN).
This concept was first introduced as Monte-Carlo (MC) Dropout in \cite{gal2015bayesian} and consequently applied to computer vision tasks such as semantic segmentation \cite{kendall2015bayesian}.
\cite{miller2018dropout} applies MC Dropout to the SSD \cite{liu2016ssd} object detection architecture.
Box detections of multiple forward passes are grouped and aggregated by Intersection-over-Union (IoU) overlap.
Consequently, uncertainties with regard to position regression and classification are derived from these samples.
\cite{kraus_uncertainty_2019} introduced Bayesian YoloV3.
In addition to MC Dropout, variances are predicted for the location parameters of the regressed bounding boxes.
\cite{harakeh2020bayesod} replaces the Non-Maximum Suppression (NMS) with bayesian inference for enhanced uncertainty estimation.
Lastly, \cite{deepshikha2021monte} proposed the use of Dropblock \cite{ghiasi2018dropblock} instead of dropout for obtaining the BNN samples.
As an alternative to MC dropout sampling, deep ensembles may be used to obtain multiple detection samples for a given image.
Deep ensembles have been proposed for OOD in image classification by \cite{lakshminarayanan2017simple}.
\cite{lyu2020probabilistic} uses ensemble samples for uncertainty estimation in object detection networks.

Sampling methods for OOD are not suitable for the use case of UAV DAA systems as multiple forward passes increase the computation time by orders of magnitude.
This renders real-time detection with high-resolution cameras infeasible on low-power hardware usually found in drones.

Sampling-free methods for OOD in object detection are considerably rarer in literature.
\cite{joseph2021towards} proposed using energy-based models on the output logits of object detection networks to discriminate between ID and OOD samples combined with a form of temperature scaling.
\cite{schubert2021metadetect} uses a post-processing method for uncertainty estimation of object detection networks.
A meta classification approach is employed that generates confidence based on box parameters and metrics derived from boxes in the NMS.

%% file: Chapters/3_model.tex
\section{Model}
\label{sec:model}
To enable OOD detection in the setting of 2D object detection with our proposed approach, only minimal changes in the loss function are required.
Particularly, it does not require any adjustments in the network architecture.
It can therefore easily be included in any modern 2D object detection framework like Faster R-CNN \cite{girshick2015fast,ren2015faster}, SSD \cite{liu2016ssd}, and the various Yolo variants \cite{redmon2016you,redmon2017yolo9000,redmon2018yolov3,ge2021yolox}.
As the architecture of the final CNN remains the same, the initial runtime of the underlying object detection framework is preserved.

\subsection{Confidence Thresholding}

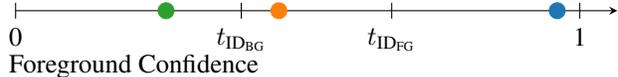
\begin{figure}[t]
  \definecolor{dblue}{rgb}{0.12156862745098,0.466666666666667,0.705882352941177}
  \definecolor{dorange}{rgb}{1,0.498039215686275,0.0549019607843137}
  \definecolor{dgreen}{rgb}{0.172549019607843,0.627450980392157,0.172549019607843}
  \centering
  \begin{tikzpicture}
    \draw [-stealth](0,0) -- (8,0);
    \draw (0,+3pt) -- (0,-3pt) node[below] {0};
    \draw (3,+3pt) -- (3,-3pt) node[below] {$t_{\text{ID}_\text{BG}}$};
    \draw (5,+3pt) -- (5,-3pt) node[below] {$t_{\text{ID}_\text{FG}}$};
    \draw (7.5,+3pt) -- (7.5,-3pt) node[below] {1};
    \node[mark size=3pt,color=dgreen] at (2,0) {\pgfuseplotmark{*}};
    \node[mark size=3pt,color=dorange] at (3.5,0) {\pgfuseplotmark{*}};
    \node[mark size=3pt,color=dblue] at (7.2,0) {\pgfuseplotmark{*}};
    \node[anchor=north west, xshift=-6pt, yshift=-13pt] at (0,0) {Foreground Confidence};
  \end{tikzpicture}
  \caption{Visualization of the thresholds over the confidence value.
    A prediction represented by the \textbf{\textcolor{dgreen}{green}} circle is treated as a background object.
    In contrast, predictions represented by \textbf{\textcolor{dorange}{orange}} and \textbf{\textcolor{dblue}{blue}} circles are both treated as \emph{actual} out-of-distribution and foreground objects, respectively.}
  \label{fig:thresholds}
\end{figure}

Multiple approaches utilize the model's raw confidence outputs computed by sigmoid or softmax activations as a base for uncertainty quantification in classification tasks \cite{schwaiger_is_2020,hendrycks_baseline_2018}.
In pure image classification, an image is either ID, \ie it has been part of the training, or OOD.
During training, the CNN is trained to best possibly recognize ID properties.
The more ID properties a CNN can detect within an image, the more confident it is with the final classification.
As a result, we expect that images with known objects, patterns, or structures have higher predicted confidence $\max_i p_i(x)$ than those with unknown ones \ie,  OOD.
Thus, it is a natural choice to distinguish between ID and OOD based on the confidence of the classification.
In object detection, however, an image may contain multiple instances of both ID -- containing both Foreground (FG) objects and Background (BG) -- and OOD objects.
To account for the presence of multiple objects and to distinguish between known foreground (ID\textsubscript{FG}), known background (ID\textsubscript{BG}), and unknown OOD objects, we impose two thresholds based on the confidence, \ie softmax score.
An illustration of the thresholds is given in \autoref{fig:thresholds}.

We can easily employ a threshold-based classifier to assign a final category $C$ of a prediction of the CNN such that
\begin{equation}
  C=
  \left\{
  \begin{array}{ll}
    \text{ID}_{\text{BG}} & \text{if } 0 \le \max_i p_i(x) < t_{\text{ID}_\text{BG}} ,                       \\
    \text{OOD}            & \text{if } t_{\text{ID}_\text{BG}} \le \max_i p_i(x)  < t_{\text{ID}_\text{FG}}, \\
    \text{ID}_{\text{FG}} & \text{if } t_{\text{ID}_\text{FG}} \le \max_i p_i(x)  < 1.
  \end{array}
  \right.
  \label{eq:func_distributions}
\end{equation}

\subsection{Margin Entropy Loss}
CNNs commonly tend to be overconfident \cite{nguyen_deep_2015}.
Therefore, predictions are also overconfident for OOD samples \cite{gal_dropout_2016}.
As a result, the distributions for both ID and OOD samples are overlapping and we can not easily separate them through the application of $t_{\text{ID}_\text{BG}}$ and $t_{\text{ID}_\text{FG}}$.
Due to this problem, we expect confidence thresholding on onchanged network outputs to not perform sufficiently well in distinguishing ID from OOD samples.
The \emph{Margin Entropy} (ME) loss term introduced in \cite{vyas2018out} as an approach to tackle the problem of overconfidence.
It has already been researched in \cite{jourdan_identication_2020} for semantic segmentation and has proven to effectively detect OOD pixels.
The ME loss is mathematically defined to be high in those cases, where the model estimates high confidence for OOD samples.
In return, it shall not affect the training of ID samples.
We, therefore, define it to be $0$ for all images with ID samples only.
As a proxy for the uncertainty of the model, we employ the Shannon entropy for discrete probability distributions $H$ with class confidences $p_i(x)$ for a given sample $x$ and $N_C$ being the total number of classes as
\begin{equation}
  H(x) = - \sum_{i = 1}^{N_C} p_i(x) \log p_i(x).
  \label{eq:entropy}
\end{equation}
A uniform distribution of the class confidences cause the entropy to be at its maximum, while in contrast, the entropy is minimal if the confidence is concentrated in a single class. We define the \emph{Margin Entropy} loss term as

\begin{equation}
  \mathcal{L}_{me} = \max{(m+\bar{H}_{\text{ID}_{\text{FG}}}-\bar{H}_{\text{OOD}}, 0)}
\end{equation}
where $\bar{H}_{\text{ID}_{\text{FG}}}$ denotes the average entropy on FG samples and $\bar{H}_\text{OOD}$ denotes the average entropy on OOD samples.
$m$ is a hyperparameter that controls the \emph{margin} between both entropies.
According to \cite{vyas2018out}, the application of a margin in contrast to entropy maximization prevents overfitting on the OOD areas and reduces negative effects on the original task.

In 2D object detection, the overall loss $\mathcal{L}$ comprises a localization loss $\mathcal{L}_\text{loc}$ and the classification loss $\mathcal{L}_\text{cls}$.
Finally, ME loss $\mathcal{L}_{\text{me}}$ is included as an additional loss such that the overall loss $\mathcal{L}$ calculates as
\begin{align}
  \mathcal{L}=\mathcal{L}_{\text{loc}}+\beta_1 \mathcal{L}_{\text{cls}} + \beta_2 \mathcal{L}_{\text{me}}.
\end{align}
$\beta_1$ and $\beta_2$ are used to control the magnitude of the loss components.

%% file: Chapters/4_experiments.tex
\section{Experiments}
\label{sec:results}

\subsection{Dataset}
In our experiments, we use a subset of the \emph{Amazon AOT Dataset} \cite{amazon_aot}, containing \num{12247} objects on \num{10071} images as training data for the network.
Additionally, we created \num{692} photorealistic renderings of \num{5} 3D models on \num{144} images for training treated as OOD samples.
Each rendered image is merged with an image from the actual Amazon AOT Dataset during training to provide a wider range of configurations.
We validated the method on a set containing \num{332} rendered OOD objects based on \num{3} different 3D models and \num{3998} ID objects on \num{3222} images.
The 3D models were exclusively used in either the training or the validation dataset.

\subsection{Experimental Setup}
\label{sec:setup}
For training and evaluation, we extend the state-of-the-art YoloX \cite{ge2021yolox} object detection framework with the ME loss. We compare the OOD detection performance to the unchanged network output trained with the standard loss function.
%
%
%
%
\subsection{Metrics}
In a typical object detection setting, metrics like Average Precision (AP) and Mean Average Precision (mAP) are used to quantify the performance of the underlying detection framework.
These metrics comprise the number of True Positive (TP), False Positive (FP), and False Negative (FP) predictions.
%
%
Besides AP, Liang \etal introduced additional metrics to assess the OOD detection performance \cite{liang_enhancing_2020}.
These metrics include FPR@\SI{95}{\percent} TPR and AUROC.
The definition of both metrics include the calculation of the False Positive Rate (FPR) defined as $\text{FPR} = \tfrac{\text{FP}} {\text{FP} + \text{TN}}$.
In particular, this metric uses the total number of True Negatives (TN).
Current state-of-the-art object detection frameworks employ a set of intial proposals called \emph{anchor} or \emph{prior boxes} \cite{liu2016ssd,redmon2016you}.
The number of those boxes scales with the image size and might reach several \num{10}k boxes per image.
As the number of \emph{real} objects, however, is typically small compared to the number of anchor or prior boxes, the number of TN will be close to the total number of those initial boxes.
This leads to $\text{FPR} \approx 0$ in most object detection frameworks even at small confidence thresholds and in particular, can change depending on the image size.
As a result, metrics that comprise the number of TN are highly influenced by the experimental setup and do not properly assess the OOD performance in object detection.

We propose to extend the confusion matrix to account for OOD objects and derive a novel \emph{Separability} metric to assess the OOD performance.

\subsubsection{Extended OOD Confusion Matrix}
The binary assignment of being either FG or BG does not specifically reflect the OOD setting as OOD objects are \emph{in between} FG and BG.
To account for those objects, we propose to add OOD as a meta-class to the confusion matrix.
This extended confusion matrix is shown in \autoref{tab:confmatrix}.

\begin{table}[t]
    \caption{Proposed confusion matrix that takes into account the OOD objects.}
    \label{tab:confmatrix}
    \centering
    \begin{tabular}{ccccc}
                                            &                                   & \multicolumn{3}{c}{\textbf{Actual}}                                                                  \\
                                            &                                   & \textbf{BG}                         & \textbf{OOD}        & \textbf{FG}                              \\\cline{3-5}
        \multirow{3}{*}{\textbf{Predicted}} & \multicolumn{1}{c|}{\textbf{BG}}  & TN                                  & FN\textsubscript{O} & \multicolumn{1}{c|}{FN}                  \\
                                            & \multicolumn{1}{c|}{\textbf{OOD}} & FO\textsubscript{N}                 & TO                  & \multicolumn{1}{c|}{FO\textsubscript{P}} \\
                                            & \multicolumn{1}{c|}{\textbf{FG}}  & FP                                  & FP\textsubscript{O} & \multicolumn{1}{c|}{TP}                  \\ \cline{3-5}
    \end{tabular}
\end{table}

We define \emph{True OOD} (TO) as actual OOD objects that have been assigned as OOD.
Furthermore, we refer to OOD objects that have been falsely assigned as FG as FP\textsubscript{O} and in return FG objects that are treated as OOD objects as \emph{False OOD/Foreground} (FO\textsubscript{P}).
Following this naming convention, we define the analogous metrics for BG objects.

\subsubsection{Separability}
To assess the quality of OOD detection within the scope of 2D object detection, we propse a novel metric called \emph{Separability} which builds upon the extended confusion matrix defined above.
Its core idea is to quantify the ability of the network to separate between FG, BG and OOD objects. It comprises two auxiliary metrics called \emph{OOD-Background Separability} (OBS) and \emph{OOD-Foreground Separability} (OFS) with each metric particularly assessing the ability to distinguish between FG or BG and OOD:
\begin{align}
    \text{OBS} & = \frac {\text{TO}} {\text{TO} + \text{FN\textsubscript{O}} + \text{FO\textsubscript{N}}}  \\
    \text{OFS} & = \frac {\text{TO}} {\text{TO} + \text{FP\textsubscript{O}} + \text{FO\textsubscript{P}}}.
\end{align}
Both OBS and OFS are $\in [0,1]$ with higher numbers representing a better ability to separate between BG and OOD, and FG and OOD.
Finally, we calculate \emph{Separability} $S(\beta)$ as
\begin{align}
    S(\beta) & = (1 + \beta^2) \cdot \frac {\text{OBS} \cdot \text{OFS}} {(\beta ^2 \cdot \text{OBS}) + \text{OFS}}.
\end{align}
Similar to OBS and OFS, $S \in [0,1]$ with higher values representing a better overall separability.
By using the parameter $\beta$ we can weight the impact of both OBS and OFS.
In our experiments, however, we use $\beta = 1$.

Contrary to the standard metrics for object detection AP and mAP, the Separability is not integrated over multiple confidence thresholds.
Instead, it assesses the model performance \emph{as is} if deployed to a productive system like autonomous drones or automated cars.
In those use cases, typically a fixed confidence threshold is chosen to allow an object to be spawned.

\subsection{Results}
We evaluate both the baseline confidences, trained with the standard cross-entropy loss, as well as the proposed method using the ME loss according to the experimental setup described in \autoref{sec:setup}.
The results are shown in \autoref{tab:results}.

%
\input{Chapters/results/99_results_table.tex}
%
The proposed ME loss approach outperforms the na\"ive baseline by \SI{129}{\percent} by means of separability $S$.
At the same time, the individual separability between OOD and BG and FG are \num{0.431} and \num{0.556}, respectively.
Thus, ME loss proves to significantly improve OOD detection in the scope of 2D object detection.

\begin{figure}[t]
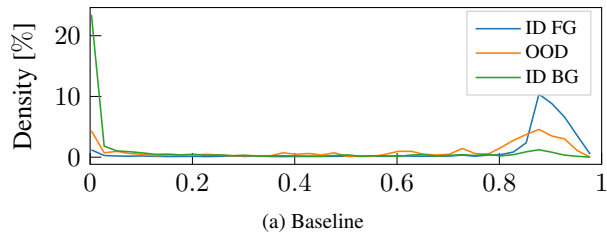
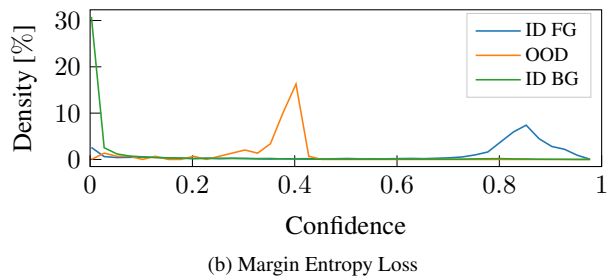

    \centering
    \subfloat[][Baseline]{\tikzBaselinePDF} \\
    \subfloat[][Margin Entropy Loss]{\tikzMElossPDF}
    \caption{Comparison of the confidence distribution after training with the baseline method (a) and the ME loss (b). We achieve a significantly better separation in (b) between all three distributions.}
    \label{fig:resDists}
\end{figure}

%% file: Chapters/results/99_results_table.tex
\begin{table}[t]
    \centering
    \caption{Results for the baseline and ME loss.
        ME loss outperforms the baseline by \SI{129}{\percent} wrt. Separability $S$.}
    \label{tab:results}
    \begin{tabular}{@{}lllllll@{}}
        \toprule
        Method           & \multicolumn{1}{c}{$S$} & \multicolumn{1}{c}{OBS} & \multicolumn{1}{c}{OFS} & \multicolumn{1}{c}{mAP@0.5} \\ \midrule
        Na\"ive Baseline & \num{0.212}             & \num{0.251}             & \num{0.184}             & \num{0.375}                 \\
        ME Loss          & \textbf{\num{0.486}}    & \textbf{\num{0.431}}    & \textbf{\num{0.556}}    & \textbf{\num{0.416}}        \\ \bottomrule
    \end{tabular}
\end{table}


%% file: Chapters/5_conclusion.tex
\section{Conclusion}
\label{sec:conclusion}
In this work, we tackled the problem of OOD detection in the scope of 2D object detection.
To this end, we proposed to extend the overall loss function by an additional Margin Entropy loss.
In addition, we introduced the Separability metric to assess the ability of a network to distinguish between Foreground, Background, and OOD.
We evaluated the proposed ME loss on the Amazon AOT dataset \cite{amazon_aot}. The ME loss outperforms the na\"ive baseline by \SI{129}{\percent} by means of Separability. Furthermore, the pure 2D object detection performance is not degraded.
The proposed loss can be included in any modern object detection framework such as Faster R-CNN \cite{girshick2015fast,ren2015faster}, SSD \cite{liu2016ssd} and the various Yolo variants \cite{redmon2016you,redmon2017yolo9000,redmon2018yolov3,ge2021yolox}.
Compared to other methods like Ensembles or MC Dropout, it does not require multiple inference per image and in return, it is able to retain the inference speed of the underlying object detection framework.
Thus, the proposed loss proves to be a suitable extension to enable OOD detection for 2D object detection in time and safety-critical applications.